\documentclass[letterpaper, 10 pt, conference]{ieeeconf}  

\IEEEoverridecommandlockouts
\usepackage{xcolor}
\usepackage{amsmath}
\usepackage{amssymb}
\usepackage{mathrsfs}
\usepackage{url}
\usepackage{hyperref}
\usepackage{mathtools}
\usepackage{subcaption}
\usepackage{xcolor}
\usepackage{algorithm}
\usepackage{algorithmic}
\usepackage{bm}
\usepackage{multirow}
\usepackage{booktabs}
\usepackage{graphicx}
\usepackage{wrapfig}

\DeclareMathOperator*{\argmin}{arg\,min}

\title{\LARGE \bf
KoopCast: Trajectory Forecasting via Koopman Operators\thanks{This work was supported in part by the Information and Communications Technology Planning and Evaluation
(IITP) grants funded by MSIT No. 2022-0-00124, No. 2022-0-00480 and No. RS-2021-II211343,
Artificial Intelligence Graduate School Program (Seoul National University).}
}

\author{Jungjin Lee \and Jaeuk Shin \and Gihwan Kim \and  Joonho Han \and Insoon Yang\thanks{All authors are with the Department of Electrical and Computer Engineering and ASRI, Seoul National University, Seoul, 08826, Korea {\tt\small \{jungbbal, sju5379, kgh3115, snowhan1021, insoonyang\}@snu.ac.kr}}%
}

\begin{document}

\maketitle
\thispagestyle{empty}
\pagestyle{empty}

\begin{abstract}
We present KoopCast, a lightweight yet efficient model for trajectory forecasting in general dynamic environments. Our approach
leverages Koopman operator theory, which enables a linear representation of nonlinear dynamics by lifting trajectories into a higher-dimensional space. The framework follows a two-stage design: first, a probabilistic neural goal estimator predicts plausible long-term targets, specifying \textit{where to go}; second, a Koopman operator-based refinement module incorporates intention and history into a nonlinear feature space, enabling linear prediction that dictates \textit{how to go}. This dual structure not only ensures strong predictive accuracy but also inherits the favorable properties of linear operators while faithfully capturing nonlinear dynamics. As a result, our model offers three key advantages: $(i)$ competitive accuracy, $(ii)$ interpretability grounded in Koopman spectral theory, and $(iii)$ low-latency deployment. We validate these benefits on ETH/UCY, the Waymo Open Motion Dataset, and nuScenes, which feature rich multi-agent interactions and map-constrained nonlinear motion. Across benchmarks, KoopCast consistently delivers high predictive accuracy together with mode-level interpretability and practical efficiency. 
\end{abstract}

\section{Introduction}

Trajectory forecasting of dynamic agents, such as pedestrians and surrounding vehicles, is a cornerstone of reliable robotic navigation systems.  
Yet, despite extensive research, forecasting remains intrinsically difficult: the intentions of dynamic agents are unobservable, their motion patterns are shaped by complex environmental contexts, and the resulting trajectories exhibit strong nonlinearity.  
To cope with these challenges, many state-of-the-art models employ large neural networks~\cite{salzmann2020trajectron++,hu2020collaborative,mao2023leapfrog,chib2024ccf}, which achieve strong benchmark performance.  
However, these approaches come with critical limitations: their black-box nature undermines interpretability and trust in safety-critical decision-making pipelines, and their substantial computational demands hinder real-time deployment in downstream planning and control.
This gap underscores the need for forecasting frameworks that combine predictive accuracy with transparency and computational efficiency, ideally by exploiting structured mathematical representations, such as linear operator models, to faithfully capture nonlinear dynamics while remaining interpretable and lightweight.

In the pursuit of interpretable and computationally efficient data-driven models for nonlinear dynamics, the Koopman operator has recently gained prominence in the robotics community~\cite{bruder2020data,mamakoukas2021derivative,haggerty2020modeling,manzoor2023vehicular}.  
This framework lifts nonlinear systems into high-dimensional linear representations, enabling data-driven methods to learn the parameters of the lifted dynamics.
By virtue of its linear structure, the Koopman operator supports fast execution and consistent long-term behavior.
Moreover, this linearity facilitates a principled analysis of the original nonlinear system, allowing it to be interpreted through standard linear algebraic tools~\cite{mezic2020spectrum}.  
Consequently, the Koopman framework offers a distinctive combination of rapid inference, strong interpretability, and competitive accuracy, making it particularly well suited for trajectory forecasting in robotic navigation.

\begin{figure*}[t]
    \centering
    \includegraphics[width=\textwidth]{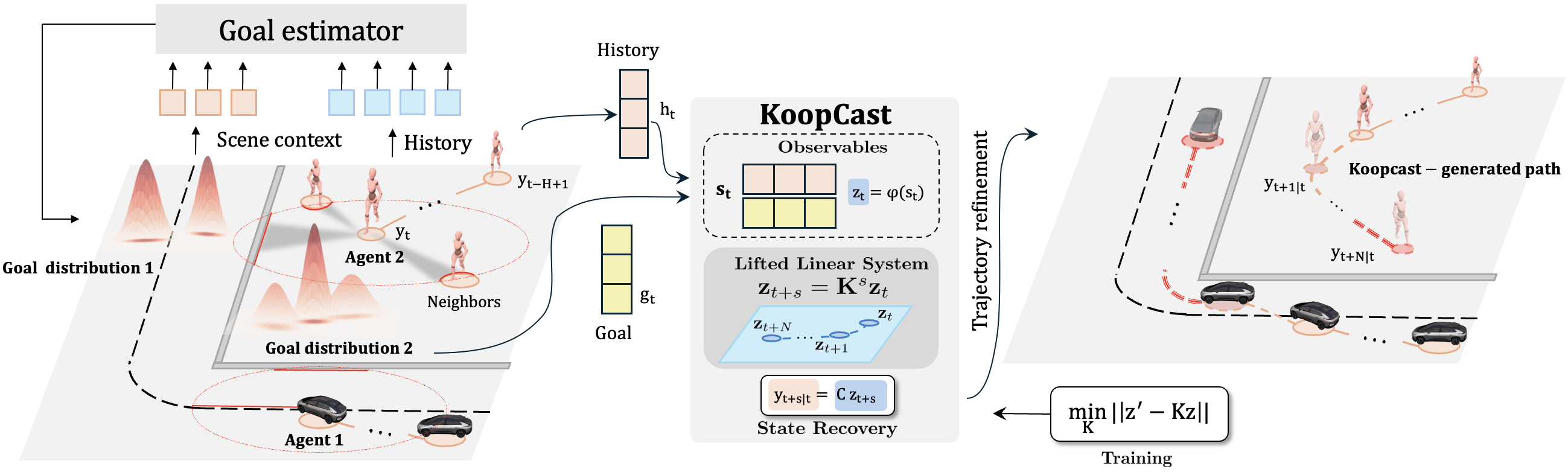}
\caption{Overview of KoopCast. The prediction process is divided into two phases: 
(1) estimating a temporal goal from history, map, and interactions, and 
(2) refining the trajectory via Koopman operator theory, where nonlinear lifting functions enhance accuracy by enabling stable and interpretable linear evolution of motion.}
    \label{fig:example}
\end{figure*}

Building on these advantages, we introduce KoopCast, a trajectory forecasting model that leverages Koopman operators in a data-driven setting.
To the best of our knowledge, KoopCast is the first model to tackle trajectory forecasting by explicitly learning the underlying Koopman operators. 
By representing nonlinear dynamics through lifted linear systems, KoopCast delivers strong performance across several dimensions: achieving high predictive accuracy, ensuring stable trajectory generation via spectral analysis, and enabling efficient training and inference through its lightweight structure.

Concretely, our approach leverages Koopman operator theory to represent dynamics as linear in a high-dimensional lifted space, incorporating both intent and history into nonlinear observables.
This is particularly powerful because, once an agent’s high-level intent (e.g., reaching a goal) is specified, its motion typically follows consistent nonlinear dynamics rather than being governed solely by uncertainty~\cite{zhao2021tnt,shi2022mtr,shi2024mtr++}.
This structure enables accurate trajectory refinement through a linear representation.
Accordingly, KoopCast performs prediction in two stages: $(i)$ goal estimation that integrates history, structural context (e.g., map and lane information), and interactions with other agents; and $(ii)$ trajectory refinement, where the Koopman operator captures nonlinear motion within a linear framework.

Through this Koopman operator-based trajectory reconstruction, our framework not only improves predictive accuracy but also enables interpretation and analysis of agent dynamics through the lens of linear algebra.
Extending this perspective, KoopCast may further contribute to a deeper understanding of pedestrian and vehicle behaviors in future research.
At the same time, it achieves low-latency inference via a simple operator-multiplication structure, making it well suited for autonomous driving applications.
Together, these properties highlight the significance of KoopCast in ensuring reliable navigation performance when integrated with downstream planners or controllers.

The contributions of this paper are summarized as follows:

\begin{itemize}
    \item We propose \emph{KoopCast}, a trajectory forecasting model grounded in Koopman operator theory to address the nonlinear nature of agent dynamics. By leveraging its linear structure, KoopCast offers a straightforward learning procedure, further enhanced by our design of goal-conditioned observable functions that improve learning efficiency.
     As a result, KoopCast consistently achieves top-3 prediction accuracy across ETH/UCY~\cite{lerner2007crowds,pellegrini2009you}, the Waymo Open Motion Dataset (WOMD)~\cite{ettinger2021large}, and nuScenes~\cite{caesar2020nuscenes}.
     
    \item We conduct spectral analysis on the learned Koopman matrices and confirm that their spectral radii remain bounded by 1, ensuring the marginal stability of forecasts.  Moreover, we identify specific Koopman modes near the unit circle ($|\lambda| \approx 1$) that dominate agent behavior, illustrating KoopCast’s potential for interpreting complex motion dynamics.  

    \item We design KoopCast with a shallow observable function and Koopman-based matrix formulation, substantially reducing computational overhead.  
Empirically, we demonstrate inference times up to 93.5\% faster than recent deep learning-based methods~\cite{salzmann2020trajectron++}, making KoopCast well suited for downstream navigation tasks. 
\end{itemize}

The implementation is publicly available at \url{https://github.com/Koopcast/Koopcast}.

\section{Related Work}

\noindent\textbf{Trajectory prediction.}
Early efforts in dynamical agent trajectory prediction (e.g., pedestrians and vehicles) employed physically grounded and rule-based models~\cite{helbing1995social,van2011reciprocal,burstedde2001simulation,schadschneider2009empirical,lefevre2014survey}.
However, these approaches suffered from limited local accuracy and an inability to capture latent features, as they relied heavily on predefined rules and prior domain knowledge.
With the rise of deep learning and large-scale datasets, data-driven methods have become dominant, spanning LSTM-based sequence models~\cite{alahi2016social,manh2018scene,altche2017lstm,deo2018convolutional}, convolutional and graph architectures~\cite{yi2016pedestrian,mohamed2020social,salzmann2020trajectron++}, generative models such as GANs~\cite{gupta2018social,sadeghian2019sophie,kosaraju2019social,hu2020collaborative} and diffusion models~\cite{mao2023leapfrog}, as well as Transformers~\cite{chib2024ccf,chib2024lg,yu2020spatio}.
Other advances include spatial reasoning through CNN-based bird’s-eye view encoders~\cite{cui2019multimodal,djuric2020uncertainty}, and recent work has even framed multi-agent forecasting as a language modeling task~\cite{seff2023motionlm}.
A particularly promising direction, highlighted in the Waymo Open Motion Challenge, is the intention–refinement paradigm~\cite{shi2022motion,shi2022mtr,shi2024mtr++,zhao2021tnt,yao2021bitrap}, where high-level intentions are first localized and then refined into detailed motions, thereby reducing multimodality and stabilizing prediction.
While these approaches achieve strong performance on large-scale benchmarks, they still face challenges such as long inference times and limited interpretability.
KoopCast addresses these limitations by providing a lightweight, supervised alternative that combines high interpretability with practical efficiency, without sacrificing predictive accuracy.

\noindent\textbf{Koopman operator theory.}
The Koopman operator~\cite{koopman1932dynamical}, a long-standing framework for representing nonlinear dynamical systems in linear form, has gained renewed interest with the rise of data-driven methods that can construct effective approximate representations~\cite{mezic2004comparison}.
Thanks to its linear structure, the Koopman framework enables the straightforward extension of control-theoretic tools developed for linear systems and has been applied to diverse robotics problems, including system identification, prediction, and model-based control~\cite{shi2024koopman,korda2018linear,zinage2023neural}.
Whereas prior efforts have primarily focused on modeling \emph{macroscopic} crowd behavior using Koopman operators (e.g.,~\cite{lehmberg2021modeling}), we instead apply the framework at the level of individual agents, directly learning and predicting their inherently nonlinear movements.
This shift allows Koopman-based methods to capture fine-grained pedestrian, vehicle, and cyclist dynamics that cannot be explained by macroscopic approximations alone.

\begin{figure*}[t]
    \centering    \includegraphics[width=\textwidth]{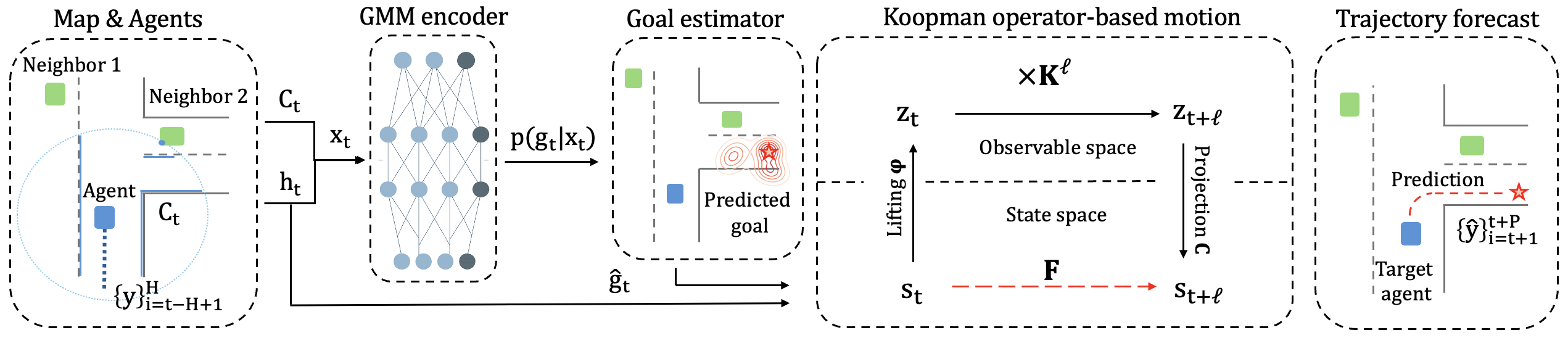}
\caption{Illustration of the KoopCast's architecture.}
    \label{fig:architecture}
\end{figure*}

\section{Background: Koopman Operators}

We consider a discrete-time nonlinear dynamical system on a state space $\mathcal{S}\subset\mathbb{R}^n$ with a nonlinear autonomous system $F:\mathcal{S}\to\mathcal{S}$. Denoting the state at step $k$ by $\mathbf{s}_k \in \mathcal{S}$, the dynamics are 
\[
    \mathbf{s}_{k+1}=F(\mathbf{s}_k), \qquad \mathbf{s}_k\in\mathcal{S}\subset\mathbb{R}^n.
\]
Let $\Phi \subset \{\, \varphi:\mathcal S \to \mathbb R \,\}$ be a vector space of \emph{observables}, for instance bounded continuous functions $C_b(\mathcal S)$ or square-integrable functions $L^2(\mathcal{S})$. The Koopman operator $\mathcal K:\Phi \to\Phi$ is defined by the composition
\begin{equation}
    (\mathcal K\varphi)(\mathbf s) \;=\; \varphi\!\big(F(\mathbf s)\big), 
    \qquad \varphi\in\Phi.
\end{equation}
The operator $\mathcal K$ is linear on $\Phi(\mathcal S)$ even when the underlying map $F$ is nonlinear. Thus, the Koopman operator linearly propagates measurement functions of the system state along the flow of the dynamics.

Ideally, one would choose observables forming an invariant subspace, but since this is infeasible, practice relies on effective observables with well-studied approximation errors~\cite{korda2018convergence}. Common choices $\{\varphi_j\}$ include polynomial basis functions~\cite{korda2018linear, bruder2020data, brunton2016koopman, kamb2020time}, approximations of Koopman eigenfunctions~\cite{korda2020optimal}, and data-driven lifting functions learned via deep neural networks~\cite{lusch2018deep, han2020deep, li2017extended}. In our work we adopt polynomial basis functions to construct the approximation.

The resulting finite-dimensional model then takes the form below.
Let $U=\operatorname{span}\{\varphi_1,\dots,\varphi_p\}\subset \Phi(\mathcal{S})$ denote an approximate $\mathcal{K}$-invariant subspace, which specifies the chosen dictionary of lifting functions. Each basis element $\varphi_j:\mathcal{S}\to\mathbb{R}$ is an observable, and the collection $\boldsymbol{\varphi}=(\varphi_1,\dots,\varphi_p)$ defines the lifting map. The lifted state
\begin{equation}
    \mathbf{z}_k=\boldsymbol{\varphi}(\mathbf{s}_k)\in\mathbb{R}^p
\end{equation}
thus lives in the finite-dimensional feature space $\mathbb{R}^p$, where the Koopman operator admits a matrix representation.
This yields the following finite-dimensional \emph{lifted linear} model
\begin{equation}\label{eq:lifted-linear}
    \mathbf{z}_{k+1} \;=\; K\,\mathbf{z}_k,
\end{equation}
where $K\in\mathbb{R}^{p\times p}$ approximates the action of $\mathcal{K}$ on $U$.  
The original state $\mathbf{s}_k$ can be recovered from the lifted coordinates through a map 
$C:\mathbb{R}^p \to \mathbb{R}^n$. Depending on the observables, the map may be as simple as coordinate extraction—as in the polynomial case where the state itself is included—or as complex as a neural network decoder.

A practical data-driven construction of $K$ is provided by extended dynamic mode decomposition (eDMD)~\cite{williams2015data}.  
Given snapshot pairs $\{(\mathbf{s}_k,\mathbf{s}_{k+1})\}_{k=1}^m$, we lift them to $\mathbf{z}_k=\boldsymbol{\varphi}(\mathbf{s}_k)$ and form
\begin{equation}\label{eq:data-matrix}
    Z \;=\; \bigl[\,\mathbf{z}_1,\ldots,\mathbf{z}_m\,\bigr],\qquad
    Z' \;=\; \bigl[\,\mathbf{z}_2,\ldots,\mathbf{z}_{m+1}\,\bigr],
\end{equation}
then solve the least-squares problem
\begin{equation}
    K \;=\; Z'\,Z^\dagger,
\end{equation}
with $Z^\dagger$ the Moore--Penrose pseudoinverse.  
Equivalently, one may use the moment matrices
\begin{equation}
    G=\tfrac{1}{m}\sum_{k=1}^m \mathbf{z}_k\mathbf{z}_k^\top,\qquad
    A=\tfrac{1}{m}\sum_{k=1}^m \mathbf{z}_k\mathbf{z}_{k+1}^\top,
\end{equation}
and set $K=G^\dagger A$. 

At prediction time, a new input $\mathbf{s}_k$ is \emph{lifted} to $\mathbf{z}_k=\boldsymbol{\varphi}(\mathbf{s}_k)$, \emph{propagated} linearly by $\mathbf{z}_{k+\ell}\approx K^\ell \mathbf{z}_k$, and then \emph{projected back} via $\hat{\mathbf{s}}_{k+\ell}=C\,\mathbf{z}_{k+\ell}$. This framework recasts nonlinear dynamics as linear evolution in a lifted space, allowing the application of linear-system tools such as control-oriented formulations, lightweight computation, and spectral analysis~\cite{mezic2020spectrum}.

\section{Trajectory Forecasting via \\ Koopman Operators}

\subsection{Problem formation}
An overview of the \emph{KoopCast} architecture is shown in Figure~\ref{fig:architecture}. We consider a point of agent with the state of $y_\tau \in \mathbb{R}^d$ ($d{=}2$ in our setting) sampled at discrete time $\tau$.
At time $t$ we observe an $H$-step history
$
h_t := (y_{t-H+1},\ldots,y_t) \in \mathbb{R}^{H\times d}
$
along with the scene context $\mathcal{C}_t = \{\, p_j \in \mathbb{R}^2 \mid \lVert p_j - y_t \rVert^2 \leq r \,\}$,  
we obtain the constrained context $c_t = \{\, p_{(j)} \in \mathcal{C}_t \mid j=1,\ldots,\min(|\mathcal{C}_t|,N) \,\}$,  
where $p_{(j)}$ denotes the $j$-th nearest point to $y_t$, i.e., we select at most $N$ nearest points to $y_t$ to reduce computational overhead while preserving locality. For pedestrian-only scenarios, the context is restricted to the
positions of other agents, whereas for vehicles, $c_t$ additionally
incorporates lane boundary and centerline points. For compactness, we denote the observed input by
$
x_t := (h_t,c_t).
$
The forecasting task is therefore to model the conditional distribution of the $P$-step future
$
y_{t+1:t+P} := (y_{t+1},\ldots,y_{t+P})$ given $x_t$.

A central challenge in future prediction lies in the high degree of uncertainty, largely stemming from the unknown intents and latent characteristics of surrounding agents. We adopted “intention/target-first” framework that localizes a destination and then refines a feasible motion around it that recent high performance models select~\cite{shi2022motion,shi2022mtr,shi2024mtr++,zhao2021tnt,yao2021bitrap}. Therefore KoopCast follows a two-stage scheme: $(i)$ estimating a distribution of temporal goals using map and agent context, and $(ii)$ refining the motion conditioned on the target.

Note that KoopCast constitutes a general framework, both theoretically and practically, as it remains largely independent of the specific agent type. Once a goal is specified, its Koopman-based refinement leverages motion history—capturing velocity, acceleration, and related dynamics—thereby ensuring consistent prediction quality across pedestrians, vehicles, and cyclists alike.

We introduce a \emph{goal} random variable at horizon $P$,
$
g \in \mathcal{G}\subseteq \mathbb{R}^d,
$
where $\mathcal{G}$ is the set of map-feasible endpoints.
Using $g_t$ as a latent variable, the future distribution can be factorized as
\begin{equation}
p\big(y_{t+1:t+P}\mid x_t\big)
=\!\!\int_{\mathcal{G}}\!\!
p(g\mid x_t)\;
p\big(y_{t+1:t+P}\mid g,x_t\big)\, dg,
\label{eq:factorization}
\end{equation}
which naturally decomposes prediction into two complementary components.
The \emph{goal predictor} estimates a compact distribution $p(g\mid x_t)$ over map-consistent temporal targets based on history and scene context, effectively capturing multimodal intent.
Conditioned on the predicted goal, the \emph{trajectory refinement module} models $p(y_{t+1:t+P}\mid g,x_t)$ through Koopman operator theory.

\subsection{Temporal goal estimator}
\label{sec:goal}
We define the temporal goal at horizon $P$ as $g_t := y_{t+P} \in \mathbb{R}^d$. To model the multimodal nature of future intentions concerning map structure and other agents' interaction, we employ a Gaussian Mixture Model (GMM) as the goal predictor. Given the observed input $x_t$—which comprises the agent’s motion history and its local scene context, i.e., $x_t = (h_t, c_t)$—the mixture density network parameterizes a Gaussian mixture distribution over $g_t$:
\begin{equation}
p(g_t \mid x_t)
= \sum_{j=1}^M \pi_j(x_t)
\mathcal{N}\big(g_t;\mu_j(x_t),\mathrm{diag}(\sigma_j^2(x_t))\big),
\label{eq:goal_mdn}
\end{equation}
where $\pi_j(x_t)$ are mixture weights satisfying $\pi_j(x_t) \geq 0$ and $\sum_{j=1}^M \pi_j(x_t)=1$, while $\mu_j(x_t)\in\mathbb{R}^d$ and $\sigma_j^2(x_t)\in\mathbb{R}^d_{+}$ denote the mean and variance parameters of each Gaussian component.
The network is trained end-to-end by minimizing the negative log-likelihood of the observed ground-truth goals, enabling it to capture diverse map-consistent endpoints. During inference, we sample multiple goals from the learned distribution, each of which leads to a distinct predicted trajectory. This empowers the model to represent the stochasticity of the prediction task. Additional architectural details of the goal predictor are provided in Appendix~\ref{app:related}.  

\subsection{Koopman operator-based trajectory refinement}

We now refine the motion behavior by generating a plausible trajectory conditioned on this estimate. Conventional trajectory refinement methods~\cite{zhao2021tnt,shi2022mtr} typically operate by progressively pruning trajectories through offset regression, iterative query updates, and candidate scoring employing either large-scaled multilayer perceptron (MLP) or transformer-based architectures. While these mechanisms can improve accuracy, they rely on dense candidate generation and multi-layer attention modules, which induce heavy computational overhead and limit interpretability.

On the other hand, KoopCast departs from heavy candidate-based refinement by leveraging a lightweight Koopman operator formulation. By lifting motion history into a higher-dimensional observable space, future trajectories evolve linearly under Koopman dynamics, with the goal providing contextual guidance. This linear propagation enables efficient closed-form rollout without requiring dense proposal sets or multi-layer attention, thus drastically reducing computational cost. Beyond efficiency, the spectral structure of the Koopman operator provides a natural lens for interpretability: eigenvalues reflect persistence or decay of dynamic modes, while eigenfunctions capture direction-specific behaviors such as turning or acceleration. As a result, KoopCast achieves both fast inference and transparent dynamics analysis, making it well-suited for downstream planning tasks where latency and reliability are critical.

Since the dominant uncertainty in trajectory forecasting arises from high-level target ambiguity~\cite{wu2023goal,liu2024uncertainty}, the refinement dynamics themselves can be reasonably modeled as \emph{deterministic} when training the Koopman operator. Extending the framework to stochastic refinement remains future work.

Building on this perspective, we reformulate trajectory prediction and demonstrate its effectiveness through empirical results. 
Let $y_t\in\mathbb{R}^d$ be the agent position at time $t$ and $h_t=(y_{t-H+1},\ldots,y_t)\in(\mathbb{R}^d)^H$ the $H$-step history. Our aim is to find the \emph{deterministic} version of trajectory refinement. In refinement, we use $(h_t,g_t)$ instead of $(h_t,c_t)$, since the context $c_t$ is already encoded in $g_t$.
Therefore, we newly construct an \emph{autonomous} augmented state by appending a temporal goal $g_t\in\mathbb{R}^d$, 
\[
\mathbf{s}_t  \;=\; (h_t,\,g_t)\;\in\;\mathcal{S}\subset\mathbb{R}^{n},\qquad n=dH+d,
\]
and posit the nonlinear but closed evolution
\begin{equation}\label{eq:augmented-autonomous}
\mathbf{s}_{t+1}=F(\mathbf{s}_t),
\end{equation} 
which shifts the history and updates the goal component. In training, we set $g_t:=y_{t+P}$ and $g_{t+1}:=y_{t+1+P}$ from data; in testing, we obtain $\hat g_t$ from a pretrained goal estimator and hold it fixed during rollout. Let $\mathcal{F}(\mathcal{S})$ be a linear space of observables and $\boldsymbol{\varphi}=(\varphi_1,\ldots,\varphi_p)^\top:\mathcal{S}\to\mathbb{R}^p$ the dictionary that defines the lifting. Ideally, the Koopman operator is defined by
\[
(\mathcal{K}\varphi)(\mathbf{s})=\varphi(F(\mathbf{s})),
\]
which acts on functions in an infinite-dimensional space.  
To obtain a computable representation, however, we restrict the attention to the 
finite-dimensional subspace
\(
U=\operatorname{span}\{\varphi_1,\ldots,\varphi_p\}
\)
and approximate \(\mathcal{K}\) by its projection onto \(U\).  
With this approximation, the lifted state
\(
\mathbf{z}_t=\boldsymbol{\varphi}(\mathbf{s}_t)\in\mathbb{R}^p
\)
evolves linearly according to
\begin{equation}\label{eq:linear-autonomous}
\mathbf{z}_{t+1}=K\mathbf{z}_t.
\end{equation}
Therefore, the nonlinear dynamics of \eqref{eq:augmented-autonomous} can be represented as a linear dynamical system~\eqref{eq:linear-autonomous} within the Koopman operator framework.

In training phase, to obtain the Koopman matrix, we used eDMD. We form snapshot pairs $\{(\mathbf{s}_t,\mathbf{s}_{t+1})\}_{t=1}^m$, lift them to $\psi_t:=\boldsymbol{\varphi}(\mathbf{s}_t)$ and $\psi_{t+1}:=\boldsymbol{\varphi}(\mathbf{s}_{t+1})$, stack
\[
\Psi=\begin{bmatrix}\psi_1^\top\\ \vdots\\ \psi_m^\top\end{bmatrix},\qquad
\Psi'=\begin{bmatrix}\psi_2^\top\\ \vdots\\ \psi_{m+1}^\top\end{bmatrix},
\]
and solve the ridge-regularized least squares
\begin{align}
K^\top \; &= \; \argmin_W \bigl\| \Psi W - \Psi' \bigr\|_F^2 + \lambda \|W\|_F^2 \label{eq:ridge}\\
          &= \; (\Psi^\top \Psi + \lambda I)^{-1} \Psi^\top \Psi'. \nonumber
\end{align}

The trajectory prediction process of KoopCast at test time is illustrated in Figure~\ref{fig:architecture}: $(i)$ The temporal goal vector $\hat{g}_t$ is computed from the history ($h_t$) and surrounding context ($C_t$) by goal predictor; $(ii)$ the estimated goal vector $\hat{g}_t$
is transformed into a lifting state, which is concatenated with the trajectory history $h_t$ to
form the full observable $z_t = \varphi(s_t)$; $(iii)$ the Koopman operator is applied to propagate the lifted
state linearly: $z_{t+1} = K z_t$; $(iv)$ the predicted physical state
$\hat{y}_{t+1|\,t}$ is extracted from the first
2 components of
$z_{t+1}$; $(v)$ This process is repeated autoregressively for $P$ steps, leveraging the same Koopman operator at each step: \begin{align}
\mathbf{z}_{t+\ell} &= K^\ell \,\mathbf{z}_t, 
\qquad \ell = 1,\ldots,P, \\
\hat{y}_{t+\ell\,|\,t} &= C\,\mathbf{z}_{t+\ell}.
\end{align}
To capture nonlinearity, we employ a second-order polynomial observable dictionary for $\boldsymbol{\varphi}$. 
The projection map $C$ then selects the first $d$ entries of 
$\mathbf{z}_{t+\ell}$ to yield the state prediction 
$\hat{y}_{t+\ell\,|\,t}$, $\ell=1,\ldots,P$.

\section{Experiments}

\subsection{Experimental setup}

\noindent\textbf{Datasets.} 
To demonstrate the efficiency of our method, we utilize several widely adopted trajectory forecasting benchmarks that provide pedestrian, vehicles, and cycles. Our benchmark datasets include pedestrian-centric datasets of ETH/UCY~\cite{lerner2007crowds,pellegrini2009you}, and large-scale autonomous driving datasets of Waymo Open Motion Dataset~\cite{ettinger2021large} and nuScenes~\cite{caesar2020nuscenes}.

\noindent\textbf{Metrics.} We follow the common practice of reporting \textit{minADE} and \textit{minFDE}~\cite{pellegrini2009you,alahi2016social}, 
i.e., best-of-$K$ displacement errors, where in our case the stochasticity comes from goal prediction. For ETH/UCY we use 8-step history and 12-step prediction, for nuScenes 2 s and 6 s (2 Hz), and for Waymo 1 s and 3 s (10 Hz).

\noindent\textbf{Baseline comparisons.} For the prediction accuracy test, we constructed the table using data reported in studies that conducted experiments under comparable conditions to ours~\cite{xu2023context,xu2022socialvae}. For the inference time, we tested \emph{Trajectron++}~\cite{salzmann2020trajectron++} and \emph{Eigentrajectory}~\cite{bae2023eigentrajectory} with the publicly available code released by those studies.

\subsection{Implementation details}

\noindent\textbf{Observable function.} To capture hidden linearity, KoopCast augments the observable dictionary with motion histories, their quadratic terms, and temporal goals, i.e., $\varphi(s_t) = [\,h_t,\; h_t^{\odot 2},\; g_t\,]^\top$. We define a projection map $\mathcal{C}:\mathbb{R}^p \to \mathbb{R}^d$ that extracts $x_t$, i.e., the leading part of the history $h_t$.

\noindent\textbf{Normalization.}  
For large-scale datasets (e.g., Waymo, nuScenes), we stabilize training by shifting the map to the origin and standardizing coordinates with training-set mean and variance.

\subsection{Main results: prediction performance}
\label{subsec:main_results_prediction}

Tables~\ref{eth_ucy_comparison}, \ref{waymo_comparison}, and \ref{nuscenes_comparison} present a quantitative comparison of KoopCast against baseline methods. Our approach achieves the most significant gains on the large-scale Waymo and nuScenes benchmarks, consistently surpassing nearly all competing predictors. On the ETH/UCY dataset, KoopCast also shows competitive results, particularly in complex and dynamic scenes (Univ, Zara01, Zara02). KoopCast attains these improvements by combining efficient closed-form rollout with interpretable spectral dynamics, yielding both fast inference and reliable performance on large-scale benchmarks.

\begin{table}[h]
\centering
\caption{Performance on ETH/UCY dataset (ADE/FDE in meters). The top three results are highlighted in bold.}
\begin{tabular}{l|c c c c}
\hline
 & ETH & Univ & Zara01 & Zara02 \\
\hline
Linear & 1.07/2.28 & 0.52/1.16 & 0.42/0.95 & 0.32/0.72 \\
SocialGAN~\cite{gupta2018social} & 0.64/1.09 & 0.56/1.18 & 0.33/0.67 & 0.31/0.64 \\
SoPhie~\cite{sadeghian2019sophie} & 0.70/1.43 & 0.54/1.24 & 0.30/0.63 & 0.38/0.78 \\
SocialWays~\cite{amirian2019social} & \textbf{0.39/0.64} & 0.55/1.31 & 0.44/0.64 & 0.51/0.92 \\
STAR~\cite{yu2020spatio} & \textbf{0.36/0.65} & 0.31/0.62 & 0.29/0.52 & 0.22/0.46 \\
TransformerTF~\cite{giuliari2021transformer} & 0.61/1.12 & 0.35/0.65 & \textbf{0.22/0.38} & \textbf{0.17/0.32} \\
MANTRA~\cite{marchetti2020mantra} & 0.48/0.88 & 0.37/0.81 & \textbf{0.22/0.38} & \textbf{0.17/0.32} \\
MemoNet~\cite{xu2022remember} & \textbf{0.40/0.61} & \textbf{0.24/0.43} & \textbf{0.18/0.32} & \textbf{0.14/0.24} \\
PECNet~\cite{mangalam2020not} & 0.54/0.87 & 0.35/0.60 & 0.22/0.39 & 0.17/0.30 \\
Trajectron++~\cite{salzmann2020trajectron++}$^*$ & 0.54/0.94 & \textbf{0.28/0.55} & 0.21/0.42 & 0.16/0.32 \\
AgentFormer~\cite{yuan2021agentformer} & \textbf{0.45/0.75} & \textbf{0.25/0.45} & \textbf{0.18/0.30} & \textbf{0.14/0.24} \\
BiTraP~\cite{yao2021bitrap} & 0.56/0.98 & \textbf{0.25/0.47} & 0.23/0.45 & 0.16/0.33 \\
SGNet-ED~\cite{wang2022stepwise} & 0.47/0.77 & 0.33/0.62 & \textbf{0.18/0.32} & \textbf{0.15/0.28} \\
\hline
KoopCast & 0.66/1.22 & 0.35/0.72 & \textbf{0.21/0.40} & \textbf{0.17/0.32} \\
\hline
\end{tabular}
\label{eth_ucy_comparison}
\end{table}

% \begin{table}[h]
% \caption{Performance on Lobby dataset}
% \label{tab:lobby}
% \centering
% \begin{tabular}{l c}
% \toprule
% \textbf{Method} & Lobby \\
% \midrule
% Trajectron++      & 0.13/0.25 \\
% Linear            & 0.30/0.59 \\
% EigenTrajectory   & 0.94/1.82 \\
% GP                & 1.81/6.85 \\
% KoopCast     & 0.12/0.22 \\
% \bottomrule
% \end{tabular}
% \end{table}

\begin{table}[h]
\centering
\caption{Performance on Waymo Open Dataset. The top two results are highlighted in bold.}
\begin{tabular}{l|c c|c c}
\hline
\multirow{2}{*}{Waymo (Validation Set)} & \multicolumn{2}{c|}{minADE$_k$ (m)} & \multicolumn{2}{c}{minFDE$_k$ (m)} \\
 & $k=1$ & $k=5$ & $k=1$ & $k=5$ \\
\hline
Constant Velocity & 2.04 & - & 5.25 & - \\
Constant Lane & 2.54 & - & 5.85 & - \\
Kalman Filter & 1.99 & - & 4.07 & - \\
SAMPP~\cite{espinoza2022deep} & 1.26 & - & 2.80 & - \\
IMAP~\cite{espinoza2022deep} & 0.97 & - & 2.03 & - \\
Trajectron++ & 0.88 & 0.56 & 2.37 & 1.41 \\
MotionCNN~\cite{konev2022motioncnn} & 0.83 & 0.40 & 1.99 & 0.81 \\
M2I~\cite{sun2022m2i} & 0.67 & 0.42 & 1.60 & 0.85 \\
ContextVAE~\cite{xu2023context}& \textbf{0.59} & \textbf{0.30} & \textbf{1.49} & \textbf{0.68} \\
\hline
KoopCast & \textbf{0.60} & 0.40 & \textbf{1.54} & 0.98 \\
\hline
\end{tabular}
\label{waymo_comparison}
\end{table}

\begin{table}[h]
\centering
\caption{Performance on nuScenes Dataset. The top two results are highlighted in bold.}
\begin{tabular}{l|c c|c c}
\hline
\multirow{2}{*}{nuScenes (Validation Set)} & \multicolumn{2}{c|}{minADE$_k$ (m)} & \multicolumn{2}{c}{minFDE$_k$ (m)} \\
 & $k=1$ & $k=5$ & $k=1$ & $k=5$ \\
\hline
Constant Velocity & 4.61 & - & 11.21 & - \\
Constant Lane & 5.45 & - & 12.73 & - \\
Kalman Filter & 4.17 & - & 10.99 & - \\
Trajectron++ & 4.08 & 2.41 & 9.67 & 5.63 \\
P2T~\cite{deo2020trajectory} & 3.82 & 1.86 & 8.95 & 4.08 \\
AutoBots-Ego~\cite{girgis2021latent} & 3.86 & 1.70 & 8.89 & 3.40 \\
ContextVAE & \textbf{3.54} & \textbf{1.59} & \textbf{8.24} & \textbf{3.28} \\
\hline
KoopCast & \textbf{2.50} & \textbf{1.73} & \textbf{4.74} & \textbf{3.31} \\
\hline
\end{tabular}
\label{nuscenes_comparison}
\end{table}

\begin{figure*}[t]
    \centering
    \includegraphics[width=0.7\textwidth]{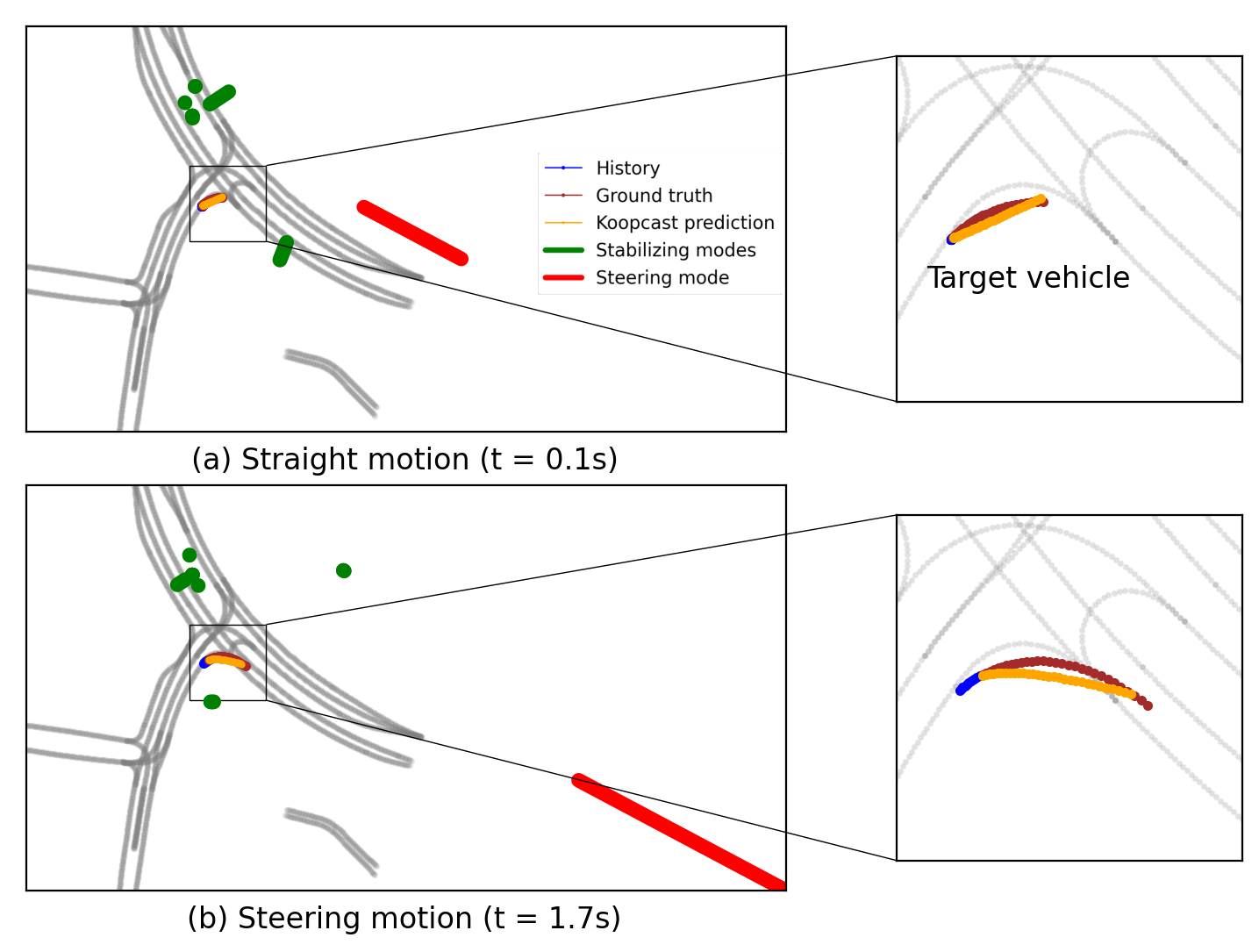}
    \caption{(a), (b) Two scenarios of eigendecomposition for motion modes. The figures at the left show the complete decomposition into all modes, while the figures right zoom into the neighborhood of the predicted trajectory. Summing all visible green and red modal contributions reconstructs the predicted trajectory (orange). (a) illustrates the mode structure while moving straight before a right turn, and the right panel shows the modes during the right turn. (b) corresponds to the initiation of a right turn. Across both cases, the steering-dominant mode (red) becomes prominently activated during turning maneuvers. Its direction aligns with the curvature of the turn, and the extended trajectory length indicates that it exerts a dominant influence on the propagation of motion, highlighting its critical role in shaping the steering behavior.}
    \label{fig:eigenmodes}
\end{figure*}

\subsection{Key properties of KoopCast}

\subsubsection{Stability guarantee}
A key requirement for trajectory refinement is stability: without it, even plausible dynamics may yield exploding or oscillatory trajectories, which is unacceptable in safety-critical forecasting. Most existing approaches address this challenge by designing elaborate refinement modules such as dense candidate selection~\cite{zhao2021tnt} or query-based iterative adjustments~\cite{shi2022mtr}. 
In contrast, our method inherently guarantees stability through the spectral properties of the Koopman operator. The lifted state $\psi_t$---which contains the physical state $x_t$ together with history and goal features---remains bounded for arbitrarily long horizons, thereby ensuring stable trajectory refinement. 
Specifically, the propagation
$\psi_{t+\ell} = K^\ell \psi_t$, 
     $\ell = 1,\ldots,P$,
remains stable when the spectral radius is no greater than 1, i.e., 
\[
    \rho(K) = \max_i |\lambda_i(K)| \;\leq\; 1,
\]
where $\{\lambda_i(K)\}_{i=1}^n$ denote the eigenvalues of $K$. 

\begin{figure}[h] 
    \centering
    \includegraphics[width=\linewidth]{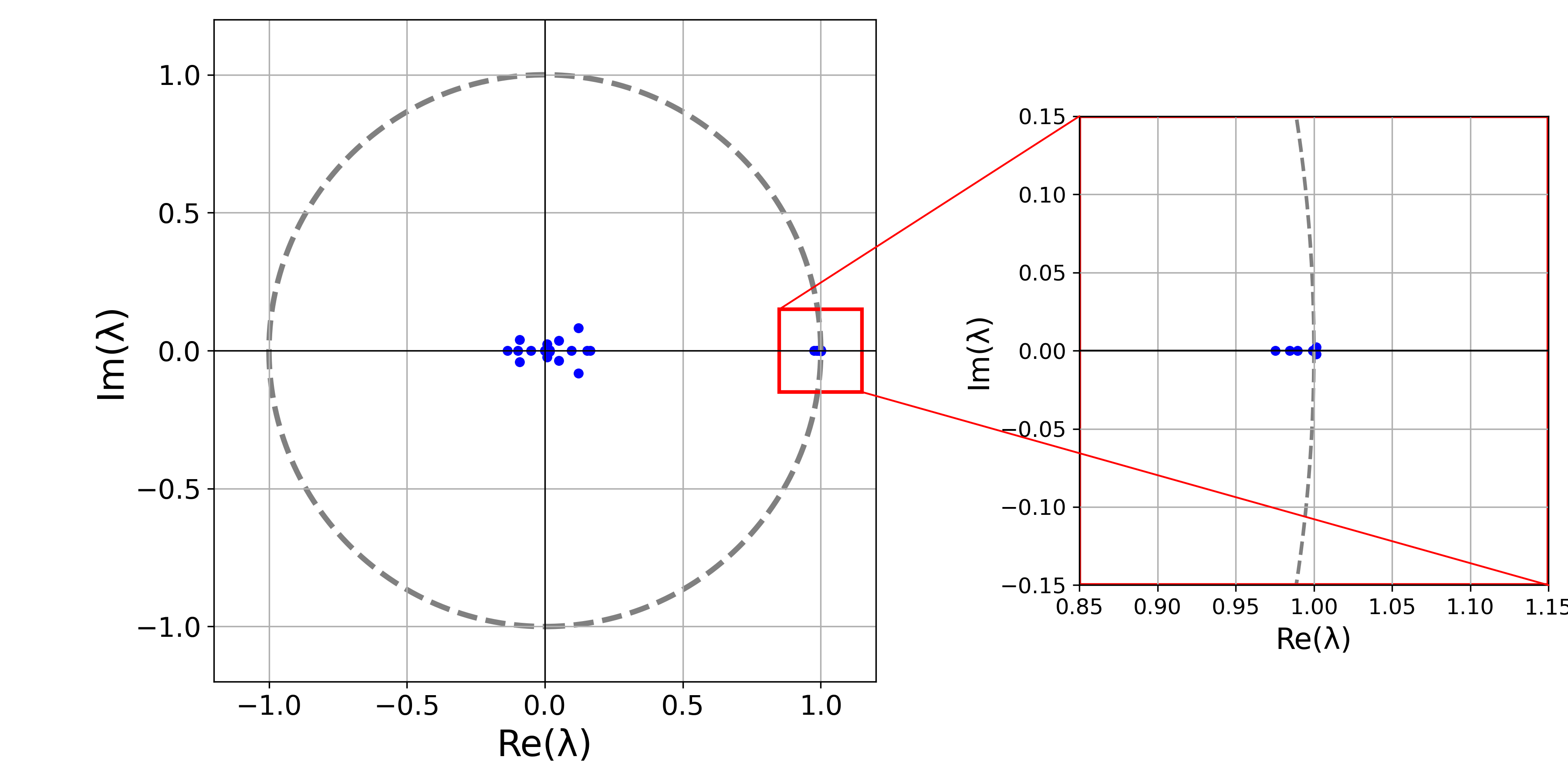}
    \caption{The spectrum of \textbf{K} scattered on the complex plane.}
    \label{fig:eigenvalue}
\end{figure}

Because the Koopman matrix $K$ is estimated via ridge regression~\eqref{eq:ridge}, its spectrum can be explicitly controlled. Figure~\ref{fig:eigenvalue} shows that, in practice, the eigenvalues cluster near or within the unit circle; moreover, increasing the ridge coefficient further contracts the spectrum, ensuring that $|\lambda_i(K)| < 1$ and thereby guaranteeing long-horizon stability.

\subsubsection{Interpretability}

Given the eigendecomposition $K = V \Lambda V^{-1}$ with eigenvalues 
$\{\lambda_i\}_{i=1}^p$ and right eigenvectors $v_i$, the Koopman rollout 
at horizon $m$ can be written as
\begin{equation}
    \hat y_{t+s} 
    = C K^m \psi_t
    = C \sum_{i=1}^p v_i \lambda_i^m \,(w_i^\top \psi_t),
    \label{eq:koopman_mode_sum}
\end{equation}
where $w_i^\top$ denotes the $i$-th row of $V^{-1}$, $\psi_t$ is the lifted state including history and goal, and $C$ projects the observable back to physical coordinates. Each term $C(v_i \lambda_i^s (w_i^\top \psi_t))$ 
corresponds to the isolated trajectory contribution of a single mode. In Figure~\ref{fig:eigenmodes} the blue and red mode curves are precisely the sequences $\{C(v_i \lambda_i^s (w_i^\top \psi_t))\}_{s=1}^P$, and the full Koopman rollout (green) is obtained by direct sum of the modes.
The modes can be broadly divided into two categories: those with $\lambda_i \geq 0.8$, which exert a dominant influence, and those with $\lambda_i \leq 0.3$, which quickly vanish. Among these persistent modes, steering maneuvers consistently activate 
a distinct \emph{steering-dominant} mode(red) whose direction aligns with 
the turn curvature.
Figure~\ref{fig:eigenmodes} illustrates this phenomenon: while all modal contributions (red/green) superimpose to reconstruct the trajectory (orange), the mode(red) is strongly amplified when the agent steers.

Beyond interpretability, this decomposition also has practical value. 
By identifying typical combinations of modes, one can flag anomalies when unusual activations appear (e.g., sudden U-turns). 
In addition, detecting strong steering-related modes allows autonomous systems to anticipate turns earlier and adjust planning in advance, improving both prediction and safety. 
A more detailed investigation of these practical applications is left for future work.

\begin{table}[h]
\centering
\caption{Average inference time per sample (ETH/UCY datasets). 
Values are averaged across all scenarios.}
\scriptsize
\begin{tabular}{l c c c c}
\toprule
 & \textbf{Linear} & \textbf{KoopCast} & \textbf{EigenTrajectory} & \textbf{Trajectron++} \\
\midrule
\textbf{Time [ms/sample]} & 0.0019 & 0.116 & 0.665 & 1.778 \\
\bottomrule
\end{tabular}
\label{tab:inference_time}
\end{table}

\subsubsection{Training and inference time}

We compare inference times on ETH/UCY in Table~\ref{tab:inference_time}. The \textit{Linear} baseline is fastest due to direct extrapolation. Excluding this trivial case, \textit{KoopCast} achieves substantially lower latency than deep models such as \textit{EigenTrajectory} and \textit{Trajectron++}, with about $93.5\%$ faster inference than the latter. This efficiency arises from its shallow goal estimator and linear eDMD refinement, avoiding costly recurrent or attention operations, and makes \textit{KoopCast} well-suited for real-time navigation and control.

\section{Conclusions}

We have presented a dynamical objects motion prediction framework grounded in Koopman operator theory, which transforms nonlinear agent dynamics into a higher-dimensional linear representation. Our predictor achieves superior performance across diverse environments containing heterogeneous dynamic agents (e.g., pedestrians, vehicles, cyclists), while ensuring stable trajectory evolution through both theoretical guarantees and empirical validation from Koopman spectral theory. Beyond accuracy, the Koopman-based formulation offers interpretability for analyzing motion dynamics, and fast training and inference.

\appendices

\section{Extended Literature Review}

\noindent\textbf{Trajectory prediction.}
Early efforts in pedestrian trajectory prediction employed physically grounded, rule-based models~\cite{helbing1995social,van2011reciprocal,burstedde2001simulation,schadschneider2009empirical}, but these approaches exhibited limited local accuracy and failed to capture latent features due to their reliance on predefined rules and prior domain knowledge. With the rise of deep learning and large-scale datasets, data-driven approaches have become dominant, spanning LSTM-based sequence models ~\cite{alahi2016social,manh2018scene}, convolutional and graph architectures~\cite{yi2016pedestrian,mohamed2020social,salzmann2020trajectron++}, generative models such as GANs ~\cite{gupta2018social,sadeghian2019sophie,kosaraju2019social,hu2020collaborative}, and diffusion models~\cite{mao2023leapfrog}, as well as Transformers~\cite{chib2024ccf,chib2024lg,yu2020spatio}, with recent work framing multi-agent forecasting as a language modeling task~\cite{seff2023motionlm}. These approaches capture multimodal and socially aware behaviors but often incur high inference costs and limited interpretability. To mitigate these issues, lightweight MLP-based models~\cite{zhang2021pedestrian,guo2023back} have been proposed, trading accuracy for efficiency.

In vehicle behavior prediction, early physics-based and probabilistic models such as kinematic filters, Hidden Markov Models, and Bayesian networks~\cite{lefevre2014survey} enabled short-horizon forecasting but proved too rigid for complex interactions, motivating deep learning approaches that capture temporal dependencies with LSTMs~\cite{altche2017lstm,deo2018convolutional}, and spatial reasoning through CNN-based bird’s-eye view encoders ~\cite{cui2019multimodal,djuric2020uncertainty}. One promising direction highlighted in the Waymo Open Motion Challenge is the intention–refinement paradigm~\cite{shi2022motion,shi2022mtr,shi2024mtr++,zhao2021tnt,yao2021bitrap}, where high-level intentions are first localized and then refined into detailed motions, thereby reducing multimodality and stabilizing prediction. These approaches have achieved strong performance on large-scale benchmarks, yet they still face challenges such as long inference times and limited interpretability. This motivates ongoing research into lightweight, and more interpretable prediction paradigms. KoopCast adopts a straightforward supervised learning framework that simultaneously offer high interpretability and computational efficiency.

\noindent\textbf{Koopman operator theory.}
The Koopman operator~\cite{koopman1932dynamical}, a long-standing framework for representing nonlinear dynamical systems as linear ones, has seen renewed interest with the rise of data-driven methods that effectively identify approximate representations~\cite{mezic2004comparison}.  
Owing to its linear nature, the Koopman framework enables the straightforward extension of control-theoretic approaches for linear systems and has found diverse applications in robotics, including system identification, prediction, and model-based control~\cite{shi2024koopman,korda2018linear,zinage2023neural}.  
In contrast to prior efforts that focus on modeling \emph{macroscopic} crowd behavior using the Koopman operator (e.g.,~\cite{lehmberg2021modeling}), we apply the Koopman framework to directly learn and predict inherently nonlinear movement of general dynamical objects (pedestrians, vehicles, cycles, etc.)

\section{Goal Estimator Design}\label{app:related}

At time $t$, we observe an $H$-step history $h_t=(y_{t-H+1},\ldots,y_t)$ and context $\mathcal{C}_t$.
The temporal goal at horizon $P$ is defined as $g_t := y_{t+P} \in \mathbb{R}^d$.

\subsection{Input preprocessing}
Before predicting the temporal goal of a target agent, say $\mathbf{A}$, we apply normalization to relevant information. Specifically, the data $x_t$ containing geometric information used to infer the goal of $\mathbf{A}$ are projected onto the $\mathbf{A}$'s ego frame. Formally, denoting the 2D pose of $\mathbf{A}$ with respect to the global frame at $t$ by $(p_t, R_t) \in \mathrm{SE}(2)$, each positional data $q_t$ (including the positions of other agents and the lane points) is transformed into
\begin{equation*}
\bar{q}_t =  R_t^{-1} (q_t - p_t).
\end{equation*}
% To normalize translation and rotation, all positions are expressed in an 
% ego-centric frame aligned with the current pose $(p_t,\psi_t) \in \mathrm{SE}(2)$. 
This transformation 
maps the current position of $\mathbf{A}$ to $(0, 0)$ and its heading to the $x$-axis.  
The transformed data $\bar{x}_t$ is now independent of the global coordinate system of the datasets. By doing so, we effectively utilize the locality of interactions and geometry. Indeed, this type of preprocessing amounts to considering the rotational and translational symmetries of the data, which enables rapid learning of the high-performance goal estimators.
The transformation also prevents the goal estimators from encountering large coordinate values, making the training of both the goal estimator and the Koopman operator numerically stable. 

% Thus, history and context features become invariant to global coordinates.

\subsection{Architecture of goal estimator}
Our goal predictor, which outputs a probability distribution over the set of two-dimensional goals $\mathcal{G}$ given an input $x_t$, is designed as a feedforward neural network. The neural network is a 2-layer MLP that outputs the parameters of a goal distribution. Specifically, for an input $x$, the final layer of the MLP outputs the parameters $\{\pi_j(x; \theta), \mu_j(x; \theta), \sigma_j(x; \theta)\}_{j = 1}^M \in \mathbb{R}^{M \times 5}$ that define the GMM $p_\theta(g|x)$:
\begin{equation*}
p_{\theta}(g| x) = \sum_{j=1}^M \pi_j(x; \theta)\,
  \mathcal{N}\!\left(g \middle| \mu_j(x; \theta),\mathrm{diag}(\sigma_j^2(x; \theta))\right),
\end{equation*}
which embraces the potential multi-modality of the goal uncertainty. The number of the modes $M$ is a hyperparameter that controls the expressiveness of the distribution $p_{\theta}(g|x)$, while choosing an excessively large $M$ may undermine both the training and inference time of the predictor. In our experiments, we used $M = 5$ or $M= 6$, which yields performance comparable to SOTA baselines while maintaining a low inference time. Other hyperparameters defining the neural network is provided in TABLE~\ref{tab:goal_pred_hparams}.
The input $x_t$ of our goal predictor is dependent of the type of agents being predicted. First of all, the motions of pedestrians tend to be less constrained by the geometry of the map. Therefore, we let the goal predictor only considers the ego-aligned history $x_t \coloneqq h_t$. On the other hand, the vehicle motions are strongly affected by the road structure, from which we make the goal predictor take both $h_t$ and nearby lane context $c_t$ extracted from $\mathcal{C}_t$ as input: $x_t \coloneqq (h_t, c_t)$. Formally, we define $
\mathcal{C}_t = \{p_j \in \mathbb{R}^2 \mid \|p_j - y_t\|_2 \leq r\}, \qquad  
c_t = \{p_{(j)} \in \mathcal{C}_t \mid j=1,\ldots,\min(|\mathcal{C}_t|,N)\},
$
where $p_{(j)}$ denotes the $j$-th nearest point to $y_t$. This restriction selects at most $N$ closest points, ensuring computational tractability while retaining the most relevant local structure.

% \subsection{Pedestrians}
% For pedestrians, whose motion is less constrained by map geometry, the predictor 
% uses only the ego-aligned history:
% \[
% p(g_t \mid h_t) 
% = \sum_{j=1}^M \pi_j(h_t)\,
%   \mathcal{N}\!\big(g_t;\mu_j(h_t),\mathrm{diag}(\sigma_j^2(h_t))\big).
% \]

% \subsection{Vehicles}
% For vehicles, road structure strongly constrains feasible futures. 
% We therefore use both ego-aligned history $h_t$ and nearby lane context $c_t$ 
% extracted from $\mathcal{C}_t$:
% Here we suppress the explicit dependence on $(h_t,\mathcal{C}_t)$ for readability, i.e.
% $\pi_j=\pi_j(h_t,\mathcal{C}_t)$ etc. Then
% \[
% p(g_t \mid h_t,\mathcal{C}_t) 
% = \sum_{j=1}^M \pi_j\,
%   \mathcal{N}(g_t;\mu_j,\operatorname{diag}(\sigma_j^2)).
% \]

\subsection{Training and inference}
Both predictors are trained by minimizing the negative log-likelihood of training data. Given a training dataset $\{ x^{(k)}, g^{(k)} \}_{k=1}^N$, the objective is given as
\begin{equation*}
\mathcal{L}_{\text{NLL}}(\theta) \coloneqq - \frac{1}{N}\sum_{k=1}^N\log p_\theta(g^{(k)}|x^{(k)}).
\end{equation*}
We employ Adam~\cite{kingma2015adam} optimizer to optimize the parameter $\theta$. Since the architecture of the model depends on the input agent type, the models are prepared and trained separately for pedestrians and vehicles.

During inference, the predictor outputs a conditional distribution $p_\theta(g \mid x)$ over goals, 
from which we draw $M$ samples 
\begin{equation*}
\{ g^{(m)} \}_{m=1}^M \sim p_\theta(\cdot \mid x).
\end{equation*}
These sampled candidates capture the multi-modality of future trajectories. For baseline comparison with best-of-$K$ Section~\ref{subsec:main_results_prediction}, we use these candidates. For downstream navigation tasks, one may alternatively use the deterministic estimate
\begin{equation*}
\bar g = \mathbb{E}_{p_\theta}[g | x],
\end{equation*}
which provides a stable goal while sacrificing diversity.

\subsection{Implementation details}
\begin{table}[t]
\centering
\caption{Hyperparameters for goal predictor. 
$^\dagger$MDN denotes Mixture Density Network.}
\label{tab:goal_pred_hparams}
\begin{tabular}{lcc}
\toprule
Hyperparam. & Waymo/nuScenes & ETH/UCY \\
\midrule
Optimizer      & \multicolumn{2}{c}{Adam~\cite{kingma2015adam}} \\
Learning rate  & \multicolumn{2}{c}{$10^{-3}$} \\
Input dim      & 98 (20+66+12) & 86 (20+66) \\
Output dim     & \multicolumn{2}{c}{674} \\
\# MDN hidden layers & \multicolumn{2}{c}{2} \\
MDN hidden units     & \multicolumn{2}{c}{128}  \\
Mixtures       & 5 & 5--6 \\
Output dim     & \multicolumn{2}{c}{2 (goal)} \\
Batch size     & 128 & single \\
Epochs         & 100 & 30 \\
\bottomrule
\end{tabular}
\end{table}

Table~\ref{tab:goal_pred_hparams} provides the hyperparameters used to train goal predictor of KoopCast across all datasets. Across ETH and UCY, we use 80\% of the trajectories for training, 10\% for validation, and
10\% for testing. For WOMD and nuScenes, 90\% of the trajectories are used for training, while the evaluation of the methods is done on their validation datasets. Our implementation relies on Pytorch~\cite{paszke2017automatic} to define and train the encoder. All
experiments are conducted on a machine equipped with an Intel Core i9-13900K @ 3.00 GHz and
an NVIDIA GeForce RTX 4090.

\bstctlcite{BSTcontrol}
\bibliographystyle{IEEEtran}
\bibliography{refs}

\end{document}